%% file: wacv2022.tex
\documentclass[10pt,twocolumn,letterpaper]{article}

\usepackage{wacv}
\usepackage{times}
\usepackage{epsfig}
\usepackage{graphicx}
\usepackage{amsmath}
\usepackage{amssymb}
\usepackage{multirow}
\usepackage{color}
\usepackage{algorithmic}
\usepackage{algorithm}
\usepackage{booktabs}
\usepackage{url}
\usepackage{soul}
\usepackage{siunitx}

\DeclareMathOperator*{\argmax}{arg\,max}

\usepackage[pagebackref=true,breaklinks=true,letterpaper=true,colorlinks,bookmarks=false]{hyperref}

\wacvfinalcopy 

\def\wacvPaperID{57} 

\begin{document}

\title{Budget aware Few-shot learning via Graph Convolutional Network}

\author{Shipeng Yan\\
ShanghaiTech University\\
{\tt\small yanshp@shanghaitech.edu.cn}
\and
Songyang Zhang\\
ShanghaiTech University\\
{\tt\small zhangsy1@shanghaitech.edu.cn}
\and
Xuming He\\
ShanghaiTech University\\
{\tt\small hexm@shanghaitech.edu.cn}
}

\maketitle

\begin{abstract}
   	This paper tackles the problem of few-shot learning, which aims to learn new visual concepts from a few examples. A common problem setting in few-shot classification assumes random sampling strategy in acquiring data labels, which is inefficient in practical applications. In this work, we introduce a new budget-aware few-shot learning problem that not only aims to learn novel object categories, but also needs to select informative examples to annotate in order to achieve data efficiency. 
   	
   We develop a meta-learning strategy for our budget-aware few-shot learning task, which jointly learns a novel data selection policy based on a Graph Convolutional Network (GCN) and an example-based few-shot classifier. Our selection policy computes a context-sensitive representation for each unlabeled data by graph message passing, which is then used to predict an informativeness score for sequential selection. We validate our method by extensive experiments on the mini-ImageNet, tiered-ImageNet and Omniglot datasets. The results show our few-shot learning strategy outperforms baselines by a sizable margin, which demonstrates the efficacy of our method. 
\end{abstract}

\input{./data/introduction}
\input{./data/relatedwork}
\input{./data/setting}
\input{./data/model}
\input{./data/experiments}
\input{./data/conclusion}


{\small
\bibliographystyle{ieee_fullname}
\bibliography{egbib}
}

\end{document}

%% file: data/introduction.tex
\section{Introduction}

Learning visual concepts with high data efficiency is a hallmark of human vision, and yet  remains a challenging task for modern deep learning-based vision systems~\cite{lake2017building}. Deep convolutional networks (ConvNets), despite its recent success in visual recognition, typically requires thousands of labeled examples from each category in a supervised learning setting~\cite{russakovsky2015imagenet}. By contrast, human visual systems are capable of recognizing a new object class from a few examples based on existing visual knowledge~\cite{lake2011one}. Achieving such data efficiency in visual learning has important impact on a broad range of real-world applications such as assistive robot~\cite{finn2017one} and medical image analysis~\cite{chung2017learning}. 

A common problem setting of such learning tasks, also known as few-shot learning~\cite{lake2011one} has attracted much attention recently, especially for the image classification task~\cite{vinyals2016matching,snell2017prototypical,finn2017model, dualattention,hao2019collect,hu2019attention}. Prior work in few-shot learning mainly focus on improving model design and/or its training strategy so that shared knowledge among similar tasks can be extracted to facilitate learning a new task. Typically, it is assumed that the few annotated examples are randomly sampled from the underlying distribution in every task.



Such data sampling strategy for novel classes, however, is rather inefficient in practical vision systems. In particular, recent study shows that human visual learning tends to select object views with a combination of high-quality and diversity~\cite{bambach2018toddler}. Moreover, many real-world applications require a costly and time-consuming procedure to manually annotate the data. Hence it is essential to choose which few examples to label for effective few-shot learning in those scenarios. 

\begin{figure}
\begin{center}
\includegraphics[width=1.0\linewidth]{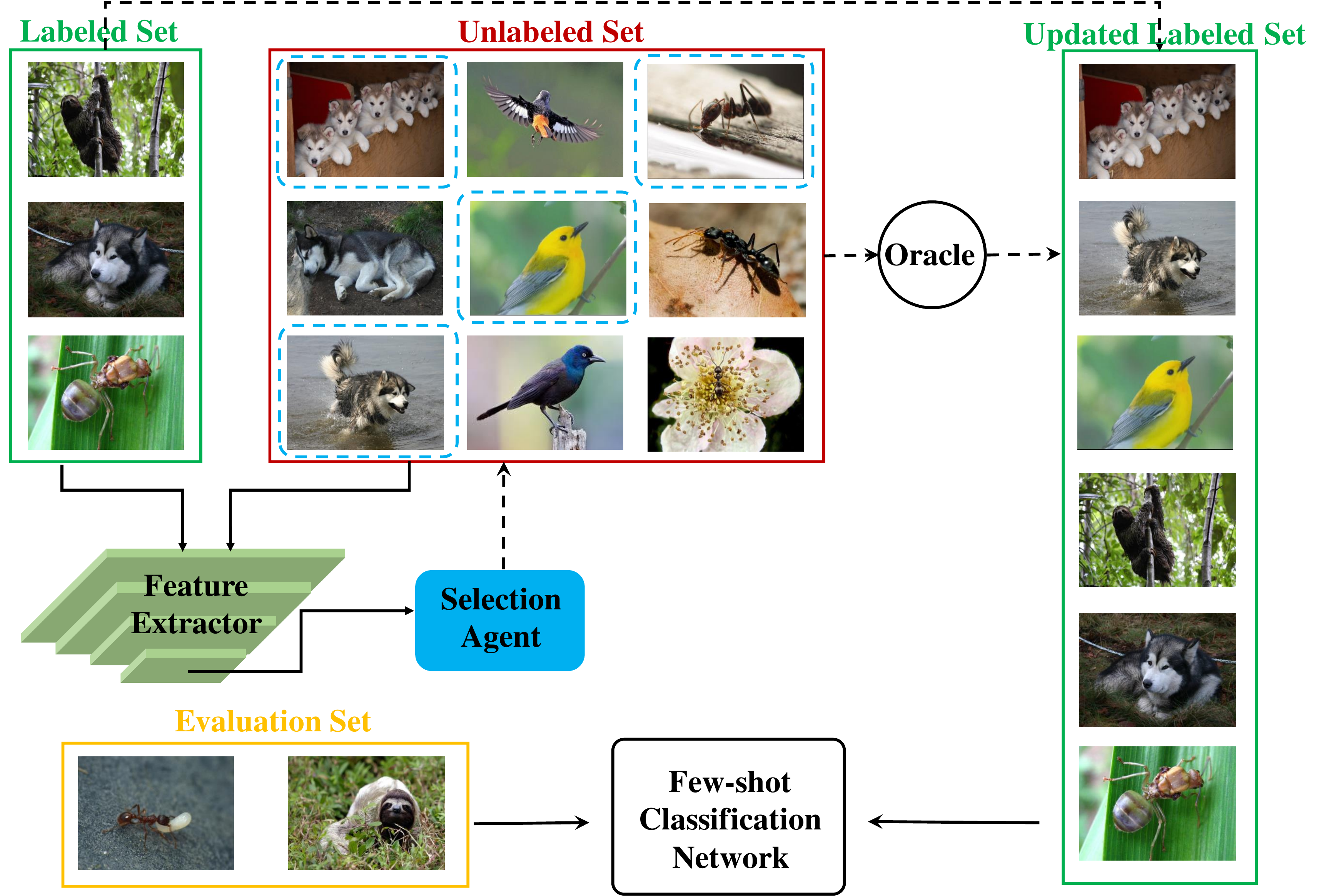}
\end{center}
   \caption{\textbf{Illustration of the few-shot learning with data selection.} First, we extract feature representations of the labeled and unlabeled instances using a feature extractor. Then the instance features and labels of the annotated data are used to select data for annotation. The instances marked with blue dotted box are selected by a learned agent. Next, we generate the updated labeled set by labeling the selected instance with an oracle. Finally, the few-shot classification is performed on the evaluation set.}\vspace{-3mm}
\label{fig:ad_figure}
\end{figure}

In this work, motivated by the above observation, we propose to tackle the few-shot learning problem by jointly solving two complementary problems, including improving data source and learning from a few training data. Our main idea is to incorporate the data annotation cost in the learning process, so that we simultaneously learn an optimized data selection policy for building or expanding the support set, and a few-shot classification strategy to predict the labels on the query set in a few-shot classification task.


To this end, we develop a meta-learning method for our active few-shot learning, which employs a novel data selection module based on a Graph Convolutional Network (GCN)~\cite{kipf2017semi} and an example-based few-shot classifier (e.g., Matching Network~\cite{vinyals2016matching}). As the classifier takes a non-parametric form, it does not require further training after the feature embedding is trained in the meta-train stage. This allows us to simplify the active few-shot learning into a task-agnostic data selection policy learning problem. 

Specifically, each few-shot learning task (referred to as an episode) includes an initial training set, an unlabeled candidate set for data selection, and an evaluation (or query) set for evaluating quality of the selection policy and the classifiers. For each task, the data selection module unrolls its policy by sequentially selecting a set of unlabeled data based on an informativeness score of each unlabeled instance. The selected data are labeled and added into the training set to build a few-shot classifier for this task. 

To avoid using hand-crafted selection criteria, we develop a graph convolutional network to learn a task-agnostic policy.  Our graph convolutional network (referred to as FL-GCN) is defined on the training and candidate instances with three sub-components for extracting distribution information of training labels, classifier outputs and data densities. We compute a context-sensitive feature representation for each unlabeled instance by propagating label and feature messages across the graph, which in turn is used to predict an informativeness score for that candidate instance. 

Our meta-training procedure samples a set of few-shot tasks from a task distribution, and iteratively updates the data selection policy based on task evaluation performances until converge. Instead of relying on reinforcement learning~\cite{pmlr-v70-bachman17a}, we adopt a supervised approach in our work to improve its efficiency. By exploiting the few-shot learning setting, we first compute an optimal ranking for unlabeled instances in each task and use gradient descent on a ranking loss of the unrolled policies for all meta-training tasks.

We conduct extensive experiment evaluations on the mini-ImageNet~\cite{vinyals2016matching}, tiered-ImageNet~\cite{ren18fewshotssl} and Omniglot~\cite{lake2011one} datasets to validate our active few-shot learning strategy. The results show that our learned policy outperforms the baselines with a large margin. We also include the detailed ablation study in order to provide a better understanding of our method.

The main contributions of this work can be summarized as the following: 
\begin{itemize}
	\setlength\itemsep{0mm}
	\item We propose a budget-aware few shot learning problem, which jointly learns a data selection policy and a few-shot classifier that optimizes the performance on novel object classes. 
	\item We develop a data selection policy network for few-shot learning that captures the distribution information of labeled and unlabeled data via a multi-component graph convolutional network. 
	\item To effectively train the task-agnostic data selection policy, we adopt a supervised learning approach and a pairwise margin ranking loss to learn the relative informativeness of the unlabeled instances.  
\end{itemize}


%% file: data/relatedwork.tex
\section{Related Work}
\paragraph{Few-shot Learning}   Few-shot learning aims to learn to recognize novel classes from only a few examples, and has recently attracted much attention~\cite{vinyals2016matching,dualattention,snell2017prototypical} since the meta-learning strategies are adopted for this task~\cite{vinyals2016matching}.
To tackle this problem, three main approaches have been proposed including metric-learning, optimization based method and parameter generation. Along the line of metric-learning,  \cite{vinyals2016matching} and \cite{snell2017prototypical} learn a cosine distance and euclidean distance that can be transferred to novel tasks, respectively. \cite{dualattention,hao2019collect,hu2019attention} introduce attention mechanism to measure the data similarity, which encodes the task-level knowledge. For optimization-based method, MAML \cite{finn2017model} tries to find better initial parameters shared by tasks so that a learner can use them to fast adapt to a new task with only few update steps. Finally, \cite{qiao2018few} propose to predict the model parameter of each novel class from the activations of a pre-trained neural network. 

In addition to image classification, few-shot learning has been extended to other vision tasks such as detection\cite{kang2019few}, semantic segmentation\cite{hu2019attention}, 3D reconstruction\cite{wallace2019few}, video-to-video synthesis\cite{wang1029fewshotvid2vid}. Moreover, \cite{boney2017semi} explores the combination of semi-supervised learning and active learning under the few-shot learning.
By contrast, we consider a new problem setting of the budget-aware few-shot learning and focus on learning a data selection policy to further improve data efficiency.

\paragraph{Active Learning} The key idea of active learning for classification tasks is to achieve the maximal accuracy by labeling the most informative data point sequentially. One of the most popular data selection strategies is the uncertainty-based method which evaluates an uncertainty score from the model prediction, such as entropy\cite{shannon1948mathematical}, least confidence\cite{settles2009active} and margin sampling\cite{scheffer2001active}. In addition, using the data distribution, \cite{settles2008analysis} introduces a strategy that estimates the information density for selecting representative data points.

Most of existing active learning methods adopt hand-crafted strategies for data selection, which have varying performances on different datasets and it is hard to determine which strategy to use for a new dataset. Motivated by this, several efforts have explored the idea of learning an active learning strategy, which has attracted increasing interests\cite{woodward2017active, pmlr-v70-bachman17a, liu2018learning, fang2017learning, NIPS2017_7010}. Most of those methods learn the strategy with deep reinforcement learning, which is inefficient and difficult to generalize. Moreover, many of them still need to use the hand-crafted uncertainty scores as the policy input. For instance, in addition to the feature representation learned by bidirectional LSTMs\cite{schuster1997bidirectional,hochreiter1997long}, \cite{pmlr-v70-bachman17a} still uses the hand-crafted feature to ensure its performance. In contrast, we propose a policy network purely based on learned representation, which does not need to use the hand-crafted rules and can be trained in a data-driven and an end-to-end manner.

\paragraph{Graph Neural Network} Recently, graph convolutional networks have been used to learn powerful feature description for the graph-structured data. The GCN is first proposed in \cite{gori2005new,scarselli2009graph,monti2017geometric}, which is the natural generalization of convolutional neural networks to non-Euclidean data. The GCN computes representations by propagating messages on a graph structure. With message traversal on the graph, representation of each node will be augmented by itself as well as other node's information. In particular, \cite{kipf2017semi} extends the convolutional networks to graphs via a localized first-order approximation and propagates the message on a dense graph composed of both labeled and unlabeled nodes. In our work, we develop a graph convolutional network with a novel message passing strategy designed specifically for the budget-aware low-shot learning problem. Unlike \cite{kipf2017semi}, our intra-and-inter message passing is able to capture the data manifold of the support set more effectively.

%% file: data/setting.tex
\section{Problem Formulation}

We aim to achieve high data efficiency in learning visual classifier by considering the few-shot image classification task. Toward this goal, we seek to jointly optimize the data sources and the visual classifiers for novel object classes. Below we introduce the setting of our budget-aware few-shot learning problem.     


The standard few-shot learning generally assumes that a few labeled examples of each novel category are sampled randomly from their underlying distributions~\cite{finn2017model}. In many realistic scenarios, however, we typically only have access to a pool of unlabeled instances and have to pay a cost to annotate them up to a budget. Hence a natural question to ask is which subset of instances we should label in order to achieve faster adaptation and better performance on the overall few-shot classification task. We refer to this new problem as \textit{the budget-aware few-shot classification}.

To address this problem, we formulate each instance of the budget-aware few-shot classification as a task (or episode) $T$ sampled from a task distribution $\mathcal{T}$. Each task is defined by a class label set $\mathbf{L}$, a support set $\mathbf{S}$  consisting of $N^l$ annotated images and $N^u$ unannotated images, an evaluation set $\mathbf{E}$ and a budget $B$. The images of support set and evaluation set are sampled from the data distributions of the classes $\mathbf{L}$. Formally, the datasets in each task are defined as follows,
	\begin{align}
	 &\mathbf{S}^{l} =\{(\mathbf{x}_{(1)}^{l}, y_{(1)}^{l}),\cdots,
						 (\mathbf{x}_{(N^l)}^{l}, y_{(N^l)}^{l})\}, \\
	 &\mathbf{S}^{u} =\{(\mathbf{x}_{(1)}^{u},\cdot),\cdots,
						 (\mathbf{x}_{(N^u)}^{u}, \cdot)\},\\
	 &\mathbf{E}=\{(\mathbf{x}_{(1)}^{e}, y_{(1)}^{e}),\cdots,(\mathbf{x}_{(N^{e})}^{e}, y_{(N^{e})}^{e})\}, 
	\end{align}
where $\mathbf{S}^l$, $\mathbf{S}^u$ denote the labeled and unlabeled training data, respectively and $\mathbf{S}=\mathbf{S}^l\cup\mathbf{S}^u$. We note that $N^l$ can be $0$, which defines a cold-start setting with $\mathbf{S}^l=\emptyset$. 

Given a task $T$, our goal is to choose a subset of $\mathbf{S}^{u}$ to label under the budget $B$ through a selection policy $\mathcal{P}$ and to build a classifier $\mathcal{C}$ on the enlarged labeled dataset that optimizes its performance on the evaluation set $\mathbf{E}$. 

\begin{algorithm}[t]
\caption{Meta-training strategy}
\begin{algorithmic}[1]
\REQUIRE 
	$p(\mathcal{T})$ is the distribution over tasks and $B$ is the budget 
\ENSURE
	$\mathcal{P}$ is the policy network.

\WHILE{not done}
	\STATE Sample a batch of tasks $T$ from task distribution
			
	\FOR{all $T=(\mathbf{L},\mathbf{S},\mathbf{E},B)$}
		\WHILE{paid cost $\le$ $B$}
			\FOR{all $u_i \in S^u_t$}
				\STATE  Build classifier $C'$ with $S^l_{aug}=\{S^l_t,u_{i}\}$ 
			    \STATE  Obtain accuracy $a^e_i$ on $E$ with $C'$
			    \STATE  Obtain instance score $s^u_i=\mathcal{P}(S_t, i)$ 
			\ENDFOR
			\STATE Obtain $\mathcal{L}_{mg}$ according to Eq.(14)
			\STATE $k = \argmax_i s^u_i$
			\STATE $S^{l}_{t+1}= \{S^l_{t},u_k \}$
			\STATE $S^{u}_{t+1}=S^u_{t} \backslash \{u_k\}$
		\ENDWHILE
	\ENDFOR
	\STATE Update $r$ using $\mathcal{L}_{mg}$
\ENDWHILE
\end{algorithmic}
\end{algorithm}

%% file: data/model.tex
\section{Our Approach}

\setlength{\abovedisplayskip}{1mm}
\setlength{\belowdisplayskip}{1mm}
\setlength{\abovedisplayshortskip}{1mm}
\setlength{\belowdisplayshortskip}{1mm}

\begin{figure*}[ht]
\begin{center}
\includegraphics[width=0.9\linewidth,height=0.161\textheight]{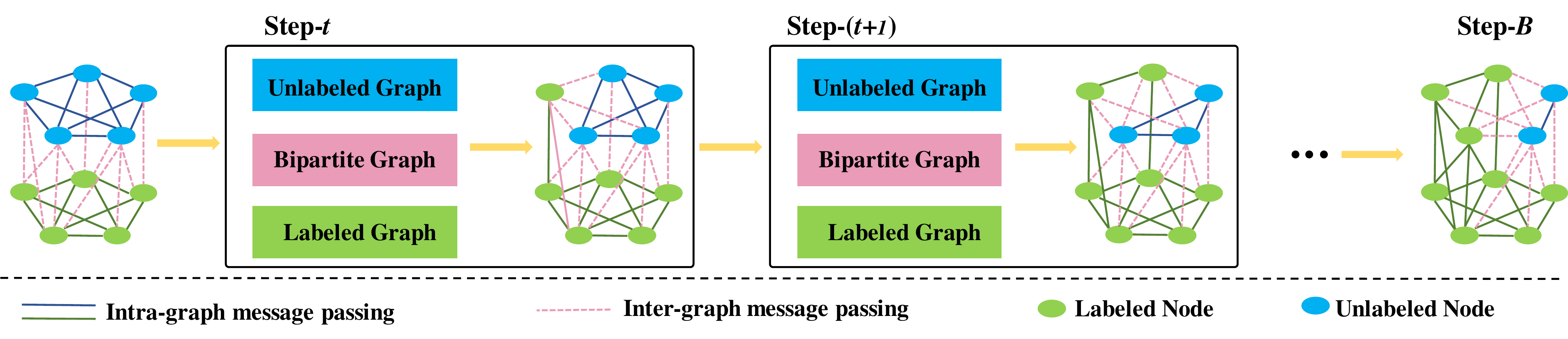}
\end{center}
\caption{An illustration of intra-graph and inter-graph message passing mechanism.}\vspace{-4mm}
\label{fig:message_passsing}
\end{figure*}

We now introduce our active learning strategy in the meta-learning framework. Solving a budget-aware few-shot learning task is challenging due to the lack of training data from novel categories and a large search space for data selection. To tackle this, we first adopt a meta-learning framework in which we learn a task-agnostic data selection policy and an example-based few-shot classifier from a set of `similar' tasks sampled from the task distribution $\mathcal{T}$. In addition, we develop a principled active learning strategy based on a learned support set representation, which sequentially chooses the most informative unlabeled instances to annotate up to the budget constraint.   


To instantiate the strategy, we introduce a deep network composed of two main modules: a \textbf{few-shot classification network} module that predicts image categories based on extracted image representations, and a \textbf{policy network} module for selecting unlabeled instances based on a graph convolutional network. 
Given a few-shot classification task, our policy network first chooses a subset of unlabeled instances to query labels and adds them into the labeled support set. The augmented support set is then utilized by the classification network to generate the label predictions for the evaluation set. 
An overview of our entire pipeline is shown in Figure \ref{fig:ad_figure}.
In this section, we first introduce the details of two network modules in Sec.~\ref{sec:embed} and Sec.~\ref{sec:policy}, followed by our meta-learning procedure in Sec.~\ref{sec:meta}.

\subsection{Few-shot classification network}\label{sec:embed}
The few-shot classification network is composed of an embedding module $\mathcal{F}$ and an few-shot classifier head $\mathcal{C}$. Given a new task $T$, our embedding module extracts the image representations of the support set, which is then used by the classifier head and the policy network in the data selection process. 

The embedding module $\mathcal{F}$ consists of a series of convolution layers with residual connections and multiple strides (e.g., a ResNet backbone network~\cite{he2016deep}). 
Concretely, for each input image $\mathbf{x}$, we denote its feature embedding as $\mathbf{z}\in \mathbb{R}^d$, and the features of the support set in a  task as $\mathbf{Z} =\{\mathbf{Z}^l,\mathbf{Z}^u\}$, where 
	\begin{align}
			&\mathbf{Z}^l = \{\mathbf{z}^l_{(1)},\mathbf{z}^l_{(2)},\cdots,\mathbf{z}^l_{(N^l)}\},\quad\mathbf{z}_{(i)}^l = \mathcal{F}(\mathbf{x}^{l}_{(i)}) \\
			&\mathbf{Z}^u = \{\mathbf{z}^u_{(1)},\mathbf{z}^u_{(2)},\cdots,\mathbf{z}^u_{(N^u)}\}, \quad  \mathbf{z}_{(i)}^u = \mathcal{F}(\mathbf{x}^{u}_{(i)})
		\end{align}
where $\mathbf{z}_{(i)}^l$ and $\mathbf{z}_{(i)}^u$ are features of labeled and unlabeled images respectively, and $d$ is the feature dimension. 

For the few-shot classifier head $\mathcal{C}$, we adopt an example-based few-shot classifier, such as the prototypical network~\cite{snell2017prototypical} for the class probability vector $p(\hat y|\mathbf{x}^u_{(i)})$ that are based on distances from the class center $\mu_k$ in the labeled set:
\begin{align}
	& \mathbf{\mu}_k = \frac{\sum_{i=1}^{N_l} \mathbf{z}^l_{(i)}\mathbf{I}[y^l_{(i)}=k]}{\sum_{i=1}^{N_l} \mathbf{I}[y^l_{(i)}=k]} \\
	& p(\hat y=k| \mathbf{x}) = \frac{e^{-d(\mathbf{z}, \mathbf{\mu}_k)}}{\sum_{j} e^{-d(\mathbf{z}, \mathbf{\mu}_{j})}} 
\end{align}
where $d$ is a distance metric(Euclidean distance is used in \cite{snell2017prototypical}) and $\mathbf{I}$ is the indicator function.

\subsection{Data Selection Policy Network}\label{sec:policy}

Given a task $T$, the data selection policy network aims to sequentially select the most informative instances from the support set $\mathbf{S}^{u}$ to annotate for building the few-shot classifier $\mathcal{C}$ so that the expected evaluation performance is optimized. Instead of using hand-crafted features and heuristics, we develop a new learnable representation that allows us to predict the informativeness of each unlabeled data instances in any task sampled from $\mathcal{T}$. Our representation is based on a graph convolutional network that describes the 
feature and label distribution, and classifier prediction uncertainty on the support set. By propagating information through the graph network, it generates a context-sensitive representation for each unlabeled example, which is then used to predict its informativeness. 

Concretely, we build a multi-component graph convolutional network, which consists of three components: \textit{1) a fully-connected labeled-data graph} $\mathcal{G}^l=(\mathcal{V}^l,\mathcal{E}^l)$, \textit{2) a fully-connected unlabeled-data graph} $\mathcal{G}^u=(\mathcal{V}^u,\mathcal{E}^u)$, \textit{3) a bipartite graph}(also called \textit{augmented graph}) $\mathcal{G}^a$ between unlabeled-data and labeled-data, to encode the overall data distribution in the support set. $\mathcal{G}^l$ and $\mathcal{G}^u$ include $N^l$ nodes $v_i^l\in \mathcal{V}^l$ and $N^u$ nodes $v_i^u \in \mathcal{V}^l$, respectively. Each node of the graph is one data point either from labeled set or unlabeled set. The augmented graph $\mathcal{G}^a=(\mathcal{V}^a, \mathcal{E}^a)$ with $N^l+N^u$ nodes, $\mathcal{V}^a=\mathcal{V}^l \cup \mathcal{V}^u$. We also introduce the dense connections between $\mathcal{V}^u$ and $\mathcal{V}^l$, denoted as $\mathcal{E}^a$. 

We introduce two sets of data representations $\mathbf{H}^l=[\mathbf{h}^l_i,\cdots,\mathbf{h}^l_{N_l}]$ and $\mathbf{H}^u=[\mathbf{h}^u_i,\cdots,\mathbf{h}^u_{N_u}]$ for $\mathcal{G}^l$ and $\mathcal{G}^u$, respectively. On each edge, we define a pairwise function $\phi(\cdot,\cdot)$ to encode the affinity between data points in the representation space. 
The representation $\mathbf{h}\in\mathbb{R}^D$ is defined as:
\begin{align}
	\mathbf{h} = [\mathbf{z};\mathbf{p};\mathbf{o}]^\intercal	
\end{align}
where $\mathbf{p}$, $\mathbf{o}$\footnote{The unlabeled data use the zero vector as the label encoding $\mathbf{o}$.} represents the classifier probability and the one-hot label encoding, respectively. In order to encode the data context from the labeled and unlabeled set, we design an iterative message propagation mechanism to update the representations in the graph network. 
%
\vspace{-3mm}
\paragraph{Message Passing Mechanism}
We propose two types of message passing mechanism in our graph convolutional network, illustrated in Figure \ref{fig:message_passsing}:

\noindent\textit{{1) Intra-graph message passing}}. We use message passing within each component graph (the labeled-data graph or unlabeled graph) to encode the global context of each subset of data instances. Take the labeled-data graph as example, we update the node representation by propagating message from each node within the labeled-data graph:
\begin{align}
	{\mathbf{\bar{h}}}^l_i = \sigma(\frac{1}{Z_{i}(\mathbf{H}^l)}\sum_{j=1}^{N_l}\phi(\mathbf{h}^l_i, \mathbf{h}^l_j){\mathbf{W}^l}\mathbf{h}_j^l)\label{eq:updating}
\end{align}
where $\bar{\mathbf{h}}^l_i \in \mathbb{R}^D$ represents the updated feature representations at node $i$, $\sigma$ is an element-wise activation function(e.g., ReLU). $Z_i(\mathbf{H}^l)$ is a normalization factor for node $i$ and $\mathbf{W}^l\in\mathbb{R}^{D\times D}$ is the learnable weight matrix defining a linear mapping to encode the message to node $i$.  

%

\vspace{1mm}
\noindent{\textit{2) Inter-graph message passing}}.
We introduce message passing between the labeled-data and unlabeled-data graph to capture the label context of unlabeled data set. With the augmented graph $\mathcal{G}^a$,
the message passing consists of updates in two directions described as follows: 
\begin{align}
\hat{\mathbf{h}}_i^l = \sigma(\sum_{j=1}^{N^u}\phi(\mathbf{h}^l_i,\mathbf{h}^u_j){\mathbf{W}^a}\mathbf{h}_j^u)	\\
\hat{\mathbf{h}}_j^u = \sigma(\sum_{i=1}^{N^l}\phi(\mathbf{h}^u_j,\mathbf{h}^l_i){\mathbf{W}^a}\mathbf{h}_i^l)	
\end{align}
where $\hat{\mathbf{h}}_i^l$ and $\hat{\mathbf{h}}_j^u$ are the updated node features for the labeled node $i$ and unlabeled node $j$, respectively. $\mathbf{W}^a\in\mathbb{R}^{D \times D}$ is the weight matrix of message parameter.

\vspace{-3mm}
\paragraph{Representation for Informativeness} 
We update the context-sensitive representations of the support set by integrating the intra-graph message $\bar{\mathbf{h}}^l, \bar{\mathbf{h}}^u$ and inter-graph message $\hat{\mathbf{h}}^l,\hat{\mathbf{h}}^u$:
\begin{align}
	\mathbf{h}^l_i=g(\left[
  		  \begin{array}{c}
  		  	\bar{\mathbf{h}}^l_i\\ 
  		  	\hat{\mathbf{h}}^l_i
  		  \end{array}
  		  \right]),\quad
  	\mathbf{h}^u_j=g(\left[
  		  \begin{array}{c}
  		  	\bar{\mathbf{h}}^u_j\\ 
  		  	\hat{\mathbf{h}}^u_j
  		  \end{array}
  		  \right])
%
\end{align}
where $\mathbf{h}^l_i$ and $\mathbf{h}^u_j$ are the updated feature representation for labeled node $i$ and unlabeled node $j$, respectively. $g$ is a feedforward fusion network. 

We can perform an iterative intra-and-inter graph message passing with multi-layer graph convolutional network, which incorporates the global context and multi-hop information for the data selection decision. For the $m$-th layer, we denote its representation as $\mathbf{H}^{(m)}$.

After the message passing and feature updating, we generate the informativeness score $s_j^u$ of each unlabeled instance by feeding the concatenation of M layer outputs $\{\mathbf{h}_j^{(m)}\}_{m=1}^M$ into a regression network $f$:
\begin{align}
	s_j^u = f([\mathbf{h}_j^{(1)},\mathbf{h}_j^{(2)}, \cdots, \mathbf{h}_j^{(M)}])
\end{align}

\subsection{Meta-Learning Strategy}\label{sec:meta}
We now detail the meta-learning strategy. Our meta-learning procedure consists of a meta-train stage with a task set $\mathbf{T}^{train}$ and a meta-test stage with its task set $\mathbf{T}^{test}$, where the image categories from two task sets are disjoint. 
\begin{figure}
\begin{center}
\includegraphics[width=1\linewidth,height=0.2\textheight]{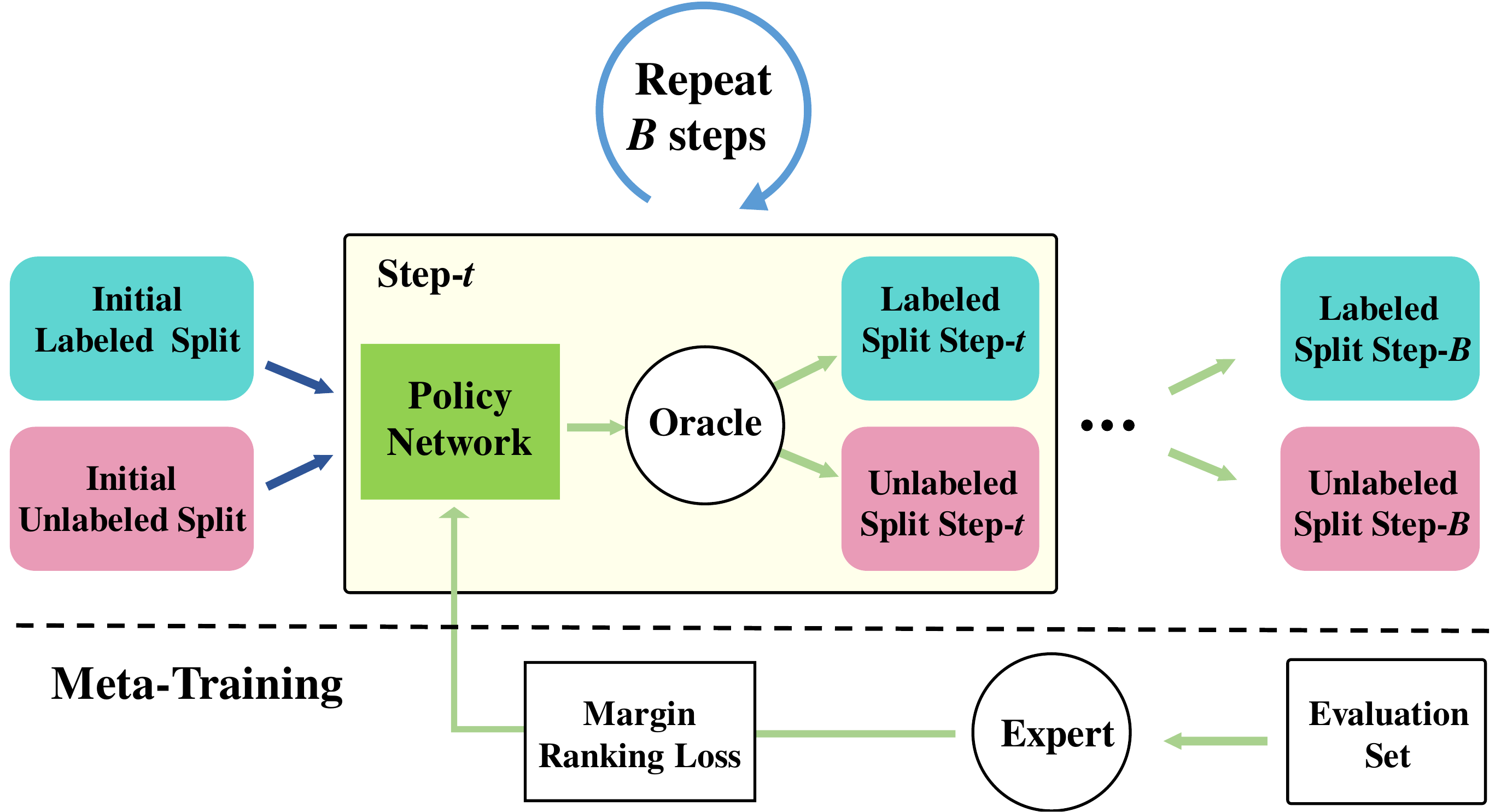}
\end{center}
\vspace{-3mm}
  \caption{\textbf{Meta-learning strategy.} The policy network is unrolled to sequentially select instances to label in the meta-test stage, corresponding to the part over the dotted lines. In the meta-train stage, the policy network is supervised by expert policy with the margin ranking loss.
}\vspace{-2mm}
\label{fig:meta_learn}
\end{figure}
In the \textit{meta-train} stage, we aim to learn the data selection policy network $\mathcal{P}$ and the few-shot classification network $\mathcal{F}+\mathcal{C}$. Our goal is to maximize the expected classification performance over the task distribution $\mathcal{T}$ under the labeling cost budget $B$. The pipeline of meta-learning strategy is shown in the Figure \ref{fig:meta_learn}.

As we employ an example-based classifier, which uses the embedding feature representation and the training set directly to make predictions, we do not need to update the meta-parameters of $\mathcal{C}$ in this work. Below we will focus on describing the learning of policy network~\footnote{We note that extending our method to other types of few-shot classifiers that requires meta-training is straightforward in the meta-learning framework, which is not our focus and left for future work.}.  

In the training process, we repeatedly sample a task from the distribution $\mathcal{T}$, execute the data selection and classification for $B$ steps, then aggregate the loss functions from each step and update the network by gradient descent. During each roll-out of the policy, the support set $\mathbf{S}=\{\mathbf{S}^l, \mathbf{S}^u\}$ will changes dynamically with $\mathbf{S}^l$ growing and $\mathbf{S}^u$ shrinking. We denote the initial state of the support set as $\mathbf{S}_0=\{\mathbf{S}^l_0, \mathbf{S}^u_0\}$, and at the $t$-th step, its state as $\mathbf{S}_{t}=\{\mathbf{S}^l_{t},\mathbf{S}^u_{t}\}$. 
 

To effectively train the policy network, we propose a supervised method by introducing an expert policy and applying dense supervision at each step $t$. 
More specifically, given a moderate-sized support set in a task, we are able to enumerate all the possible selections at each step $t$, and use the performances of resulting classifiers on evaluation set $\mathcal{E}$ to determine which unlabeled instance is most informative. This expert policy is used to optimize the policy network for data selection strategy in an imitation learning manner.

While the previous work on active learning attempts to regress the error reduction~\cite{NIPS2017_7010, settles2009active}, we find it leads to unstable training process. Instead, we adopt a new loss function inspired by the margin ranking loss~\cite{vo2016localizing}. Concretely, we propose to learn the ordering of any pair of instances in the unlabeled set by defining the following soft-margin ranking loss at each time step:
\begin{align}
\mathcal{L}_{mg}(\mathbf{S}_t) = \sum_{i<j}^{|S_t|} \max \{0, 1-\epsilon(a^e_i-a^e_j)(s^u_i-s^u_j)\}
\end{align}
where $|S_t|$ is denoted as the cardinality of $S_t$, and $s^{u}_i, s^{u}_j$ are the score of instance $i$, $j$ predicted by the policy network $\mathcal{P}$ respectively; $a^e_i, a^e_j$ represents the accuracy on the evaluation set if the unlabeled instances $i$, $j$ are added into the labeled set $\mathbf{S}^l$ respectively, and $\epsilon$ denotes the sign function.

The overall objective function for training the policy network $\mathcal{P}$ can be written as:
\begin{align}
	\mathcal{L} = \mathbf{E}_{T \sim \mathcal{T}} \left[ \frac{1}{B} \sum_{t=1}^{B} \mathcal{L}_{mg}(\mathbf{S}_t)\right]
\end{align}

%% file: data/experiments.tex
\section{Experiments and Analysis}
\begin{table*}[t]
\caption{The performance with 95\% confidence of  data selection policies on mini-ImageNet.(\textbf{CI}:Class Information)}
\small
\center
\resizebox{0.85\textwidth}{!}{
\begin{tabular}{lcccc|cc}
\toprule[0.3mm]
\multicolumn{1}{c}{\multirow{2}{*}{Methods}} & \multicolumn{4}{c}{Cold Start}                                & \multicolumn{2}{c}{Warm Start}   \\ \cline{2-7} 
\multicolumn{1}{c}{}                         & 5-way Accuracy & \#Avg Class & 10-way Accuracy & \#Avg Class & 5-way Accuracy & 10-way Accuracy \\ \hline
Expert    &    $71.90\pm0.29\%$           &  $4.73\pm0.00$           &    $59.13\pm0.14\%$             &    $8.63\pm0.01$          &     $82.70\pm0.29\%$            &     $70.73\pm0.25\%$   \\         
Random+CI & $53.77\pm1.23\%$ &  $5.00\pm0.00$ & $36.92\pm1.22\%$ & $10.00\pm0.00$ & $67.15\pm0.49\%$    &  $51.53\pm0.49\%$   \\ \hline
Random   &  $45.60\pm0.47\%$ &  $3.46\pm0.01$ &  $32.68\pm0.21\%$  &  $6.69\pm0.05$   &  $66.80\pm0.48\%$ & $51.87\pm0.18\%$ \\
Entropy  & - &   - &  -  & - &   $67.21\pm0.31\%$  &    $50.74\pm0.39\%$ \\
Min-Max-Cos &   $50.98\pm0.60\%$  &  $3.89\pm0.01$ & $35.87\pm0.19\%$ & $7.23\pm0.01$  & $66.40\pm0.49\%$ &  $52.03\pm0.36\%$ \\
K-center greedy  &  $51.29\pm0.51\%$  & $3.90\pm0.03$ & $35.59\pm0.32\%$ & $7.24\pm0.04$ & $67.12\pm0.30\%$ & $51.37\pm0.38\%$ \\
\textbf{FL-GCN} &  $\mathbf{54.00\pm0.64\%}$ & $\mathbf{4.05\pm0.04}$ & $\mathbf{38.52\pm0.53\%}$ & $\mathbf{7.37\pm0.07}$ & $\mathbf{69.14\pm0.78\%}$ &  $\mathbf{53.45\pm0.33\%}$            \\
\bottomrule[0.3mm]
\end{tabular}
\label{tab:miniImgNet}}
\end{table*}
To validate our method for the budget-aware few-shot classification problem, we conduct extensive experiments on mini-ImageNet~\cite{vinyals2016matching}, tiered-ImageNet~\cite{ren18fewshotssl} and Omniglot~\cite{lake2011one}  with detailed ablation study. We report the experiment setup in Sec.~\ref{subsec:setup}, followed by the introduction of comparison methods in Sec.~\ref{subsec:baseline}. Then we present our experiment analysis on the mini-ImageNet dataset in Sec.~\ref{subsec:mini-ImageNet}, followed by our results on both tiered-ImageNet and Omniglot datasets in Sec.~\ref{subsec:tiered_and_omniglot}. In the end, we detail the ablation study and analysis in the Sec.~\ref{subsec:ablation}.


\subsection{Experiment Setup}\label{subsec:setup}
We perform different $N$-way $K$-shot active learning task experiments.  More specifically, $N$ novel classes are sampled in one task, and the labeled set will be enlarged from the initial size to budget $B=NK$ labeled instances by querying with the policy network. 

To validate our proposed method effective for the real applicable scenarios, we conduct two kinds of experiment setting for comprehensive evaluation, which include: \textit{1) cold-start  few-shot learning} and \textit{2) warm-start  few-shot learning}. The cold-start setting is a generic setting where the labeled set starts with an empty set and continuously enlarges with newly added instances, thus requires the policy network be able to identify the missing classes without any information of the novel classes. In addition, we also explore the warm-start setting where the labeled set begins with one labeled instance per class. Compared with the cold-start setting, the warm-start setting evaluates the capability of the policy network on finding the most informative instance for an existing classifier. We report the performance when the paid cost reaches budget. Moreover, the average number of classes in the selected subset is also reported in the cold-start setting to evaluate the ability to discriminate different classes.

\subsection{Comparison Methods}\label{subsec:baseline}
For comparison, we choose several common methods as the baseline: \textit{1) Random} selects the instance in the unlabeled set with uniform probability.  \textit{2) K-center greedy\cite{sener2017active}} wants to select a subset of points as centers to minimize the largest distance between a data point and its nearest center, and optimizes it in a greedy method. \textit{3) Entropy\cite{shannon1948mathematical}} is a popular uncertainty-based active learning strategy, which is defined as follows:
\begin{align}
\sum_{c=1}^{N} -p(y=c)\log p(y=c)
\end{align} where N is the number of classes. The instance with maximum entropy will be  selected. \textit{4) Min-Max-Cos\cite{pmlr-v70-bachman17a}} is a heuristic policy which select the image has minimum maximum cosine similarity to labeled set images in the feature space. 

It's worth noting that we also build an \textit{\textbf{Expert}} method, where the score of an instance is defined as the accuracy of classifier with the updated labeled set on evaluation set if the instance is labeled and added into the labeled set. It serves as both the upper bound and the expert for policy network to imitate.

\subsection{mini-ImageNet}\label{subsec:mini-ImageNet}
 \paragraph{Dataset.} mini-ImageNet, proposed by ~\cite{vinyals2016matching}, consists of images from 100 different classes with 600 examples per class. Specifically, we adopts the splits proposed by~\cite{ravi2016optimization} with 64 classes for training, 16 for validation and 20 for testing in the meta-learning setting. Moreover, the images are resized into $84\times84$ pixels.

\paragraph{Implementation Details} We use a modified version of ResNet-10, proposed in ~\cite{mishra2018a}, and adopt prototypical network as the few-shot classifier. The graph convolutional network used is 2-layer, and the regression network $f$ is a two-layer MLP with 32 hidden units and using LeakyReLU as activation function in its hidden layer. We adopt the cosine distance as the function $\phi$ to measure the similarity. In cold-start setting, we replace the classifier probability with the min, mean and max probabilities to handle the missing class problem. For each episode, the initial unlabeled and  evaluation set are built by sampling 10 images per class where the images in both sets are disjoint. To be fair to compare with other baselines, we pretrain the few-shot classifier on the meta-train split and fix the representation. Then we apply the ADAM\cite{kingma2014adam} optimizer to train the policy network for 2000 iterations with learning rate 3e-4, weight decay 5e-5 and batch size 16.
\vspace{-5mm}
\paragraph{Quantitative Results.}
In the Table~\ref{tab:miniImgNet}, we compare our proposed method with other baselines for both cold-start and warm-start setting. 

In the cold start setting, we compare our model with the baselines in two types of situations. We only consider $N$-way 1-shot setting here. The first type is allowing the policy to select data arbitrary at each time step without any constraints. And the second type is requiring the policy to only choose instances from the different classes at each time step, which guarantees all novel classes exist in the labeled set in the ending. The second type can be considered as adding the class information into the selection policy, which is denoted as \textbf{CI} in the results table.

As shown in the Table \ref{tab:miniImgNet}, the accuracy and the average number of classes for our approach outperforms consistently all baselines by a large margin at the first situation, in both 5-way and 10-way setting. The superiority of our method on the average number of classes shows that our model has better ability to identify missing classes. Although the class information is added into the random policy(Random+CI) in the second situation, our method(FL-GCN) is still superior to the (Random+CI) which shows that our policy is not only able to distinguish different classes but can select the representative points.


We also demonstrate our method also works well and outperforms baseline methods by a sizable margin in warm-start settings. The results in the Table.~\ref{tab:miniImgNet} shows that the handcrafted strategies can not perform better than random strategy consistently which shows that the success of handcrafted strategies requires some dataset characteristics to satisfy. In summary, the above results prove that it is an effective way to improve the efficiency of data selection policy by leveraging the related active learning experiences.

\subsection{tiered-ImageNet and Omniglot}\label{subsec:tiered_and_omniglot}
\paragraph{Dataset.} We also validate our method on the tiered-ImageNet and Omniglot which has  larger and smaller difference between train and test split compared with mini-ImageNet, respectively. For tiered-ImageNet, it is proposed by \cite{ren18fewshotssl}, and is a larger subset of ILSVRC-2012\cite{russakovsky2015imagenet}. Different from mini-ImageNet, tiered-ImageNet simulates a much more difficult scenario where the high-level categories are disjoint between training, validation and test. Specifically, we follow the split proposed by \cite{ren2018incremental} with 151 classes for meta-train, 97 for meta-val and 160 for meta-test. For Omniglot~\cite{lake2011one}, the dataset consists of 1623 characters(classes) from multiple alphabet vocabularies. We follow the setting in \cite{vinyals2016matching} to split the dataset into 1200 classes for training and the remaining 423 classes for testing.
The Omniglot dataset has been widely used for testing few-shot learning methods, and recent methods achieve strong performances. Here we use it as a sanity check to validate our method.

\begin{figure*}[ht]
	\vspace{-55pt}
	\begin{minipage}[c]{0.5\linewidth}
		\centering
		\includegraphics[width=0.85\linewidth,height=0.16\textheight]{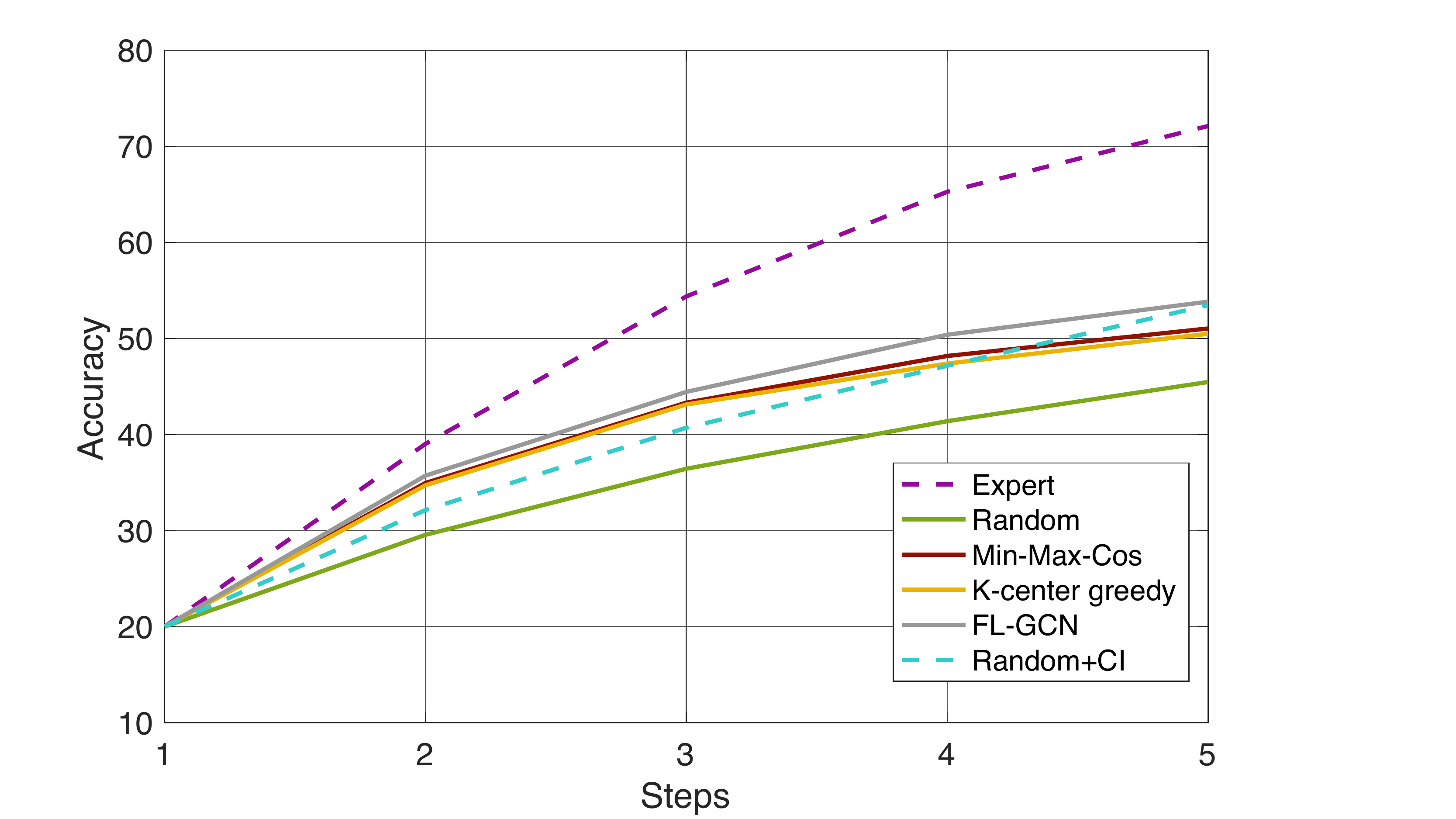}
	\end{minipage}
	\begin{minipage}[c]{0.5\linewidth}
		\centering
		\includegraphics[width=0.8\linewidth,height=0.35\textheight]{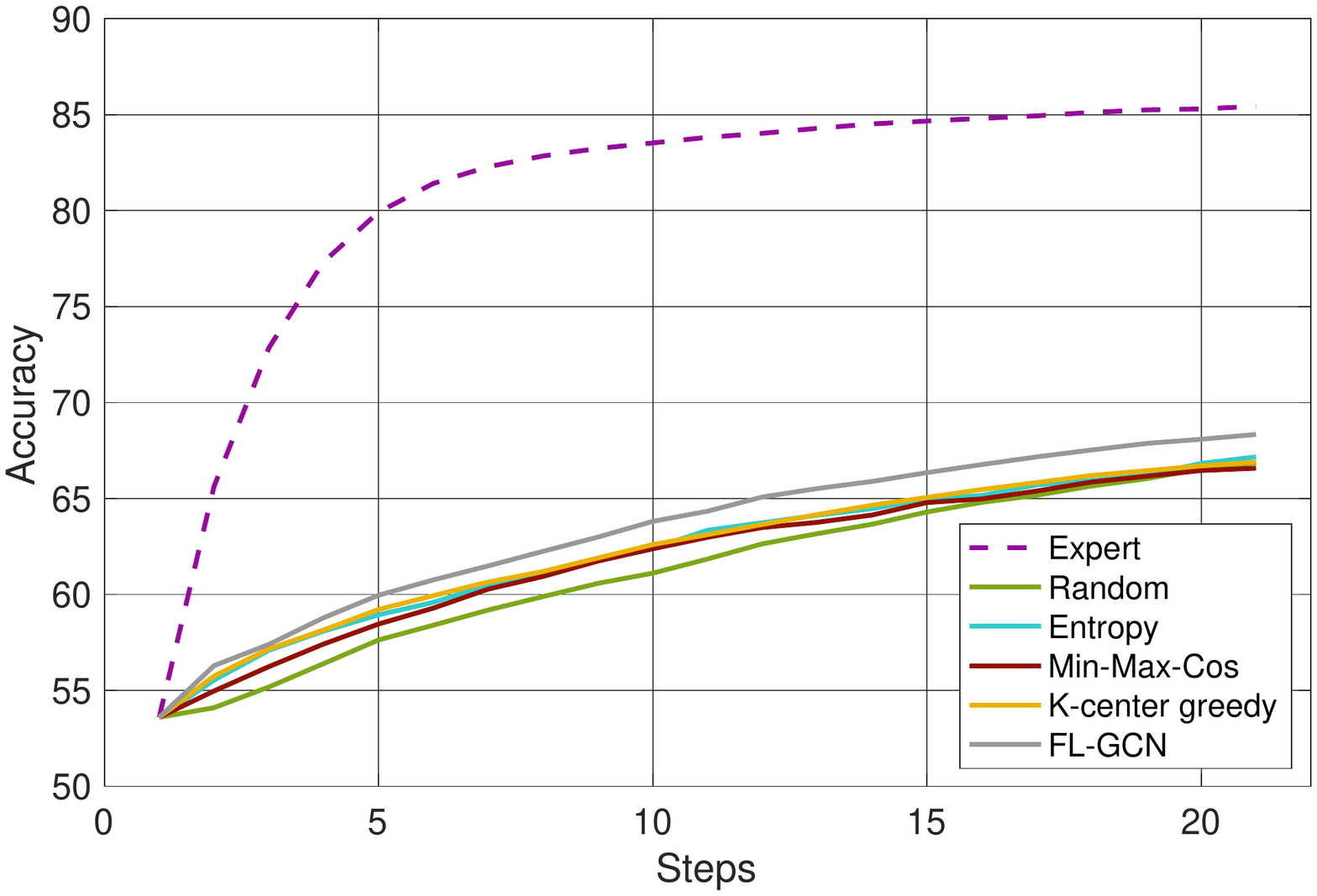}
	\end{minipage}%
	\vspace{-45pt}
	\caption{\textbf{Left}: The performance at each step in 5-way 1-shot cold-start setting. \textbf{Right}: The performance at each step in 5-way 5-shot warm-start setting.}
	\label{fig:anytime}
\end{figure*}
\vspace{-3mm}
\paragraph{Implementation Details} For tiered-ImageNet, we adopt the standard ResNet-18 and the Prototypical network as the backbone and the few-shot classifier, respectively. For Omniglot, we use ConvNet64~\cite{vinyals2016matching} which is composed of four convolutional blocks and the matching network\cite{vinyals2016matching} as the backbone and the few-shot classifier, respectively. Following \cite{santoro2016meta}, we resize the image into $28 \times 28$ pixels and augment the dataset by rotations in multiple of $\ang{90}$.

\begin{table}[t]
	\small
	\center
	\caption{The tiered-ImageNet performance with 95\% confidence of data selection policy on both 5-way 1-shot cold-start(CS) and 5-way 5-shot warm-start(WS) settings.}
	\resizebox{0.42\textwidth}{!}{
	\centering
	\begin{tabular}{cccc}
	\toprule[0.3mm]
	Method & 5-way 1shot(CS) & Avg of cls & 5-way 5shot(WS) \\
	\midrule
	Expert & $71.26\pm0.14\%$ & $4.77\pm0.01$ &  $83.46\pm0.19\%$\\
	Random & $43.17\pm0.11\%$ & $3.44\pm0.01$  & $67.21\pm1.47\%$ \\
	K-center greedy & $49.32\pm0.21\%$ & $3.91\pm0.01$ & $67.67\pm1.75\%$\\
	Min-Max-Cos & $48.87\pm0.10$ & $3.82\pm0.01$ & $67.25\pm0.58\%$\\
	Entropy & - & - & $67.60\pm0.14\%$ \\
	\textbf{FL-GCN} & $50.39\pm0.45\%$ &  $3.98\pm0.01$ & $68.86\pm0.50\%$ \\
 	\bottomrule[0.3mm]
	\end{tabular}
	\label{tab:tiered_imgnet}}
	\vspace{-3mm}

\end{table}

\vspace{-3mm}
\paragraph{Quantitative Results} 
For tiered-ImageNet, we compare the performance of our proposed FL-GCN with the other methods using the same embedding network in Table.~\ref{tab:tiered_imgnet}. We can see that our methods outperforms the baseline methods in both settings. For Omniglot, the overall comparison results are shown in the Table.~\ref{tab:omniglot}.  LAAL\cite{pmlr-v70-bachman17a} is a bi-directional LSTM based policy network. To be fair with comparison, we report the relative gain based on the performance of random strategy, which shows that our method achieves maximum gain. The results of all three datasets illustrates that our method is robust to the distance between meta-train and meta-test and have strong ability to transfer.


\begin{table}[t]
	\small
	\center
	\caption{The Omniglot performance with 95\% confidence of data selection policy on 5-way 1-shot cold-start(CS) setting.}
	\resizebox{0.37\textwidth}{!}{
	\centering
	\begin{tabular}{ccc|cc}
	\toprule[0.3mm]
	Method & 5-way 1shot(CS) & Avg \# of cls & 10-way 1shot & Avg \# of cls \\
	 \midrule
	 Random & $68.47\pm0.39\%$ &  $3.45\pm0.02$ & $65.71\pm0.16\%$ & $$ \\ \hline
	 Expert & $99.39\pm0.03\%$ & $5.00\pm0.00$ & $99.5\pm0.01\%$ & $10.00\pm0.00$ \\
	 K-center & $95.43\pm0.10\%$ & $4.87\pm0.00$ & $93.23\pm0.20\%$ & $9.55\pm0.03$ \\
	 LAAL\cite{pmlr-v70-bachman17a} & $97.4\pm0.11\%$ & - & $94.3\pm0.24\%$ & -\\
	\textbf{FL-GCN} & $96.35\%0.12$ & $4.89\pm0.01$ & $93.55\pm0.20\%$ & $9.56\pm0.02$ \\
 	\bottomrule[0.3mm]
	\end{tabular}
	\label{tab:omniglot}}
\end{table}



\subsection{Ablation study and analysis}\label{subsec:ablation}
We conduct a series of ablation study to evaluate our method from different perspectives including: \textit{1)model structure}, \textit{2) training loss} \textit{3) anytime performance}

\begin{table}[t]
	\small
	\center
	\caption{Ablation study (\textbf{Avg \# of Cls}: The average number of classes of labeled set in the ending in cold-start 1-shot setting; \textbf{FCGCN}: Fully Connected GCN where the message is passed across the single fully connected graph; \textbf{MRL}: margin ranking loss; \textbf{MSE}: Mean squared loss where the policy network is to regress the error reduction. And the subscript denotes as the number of layers)}
	\resizebox{0.5\textwidth}{!}{
	\begin{tabular}{cccc|c}
	\toprule[0.3mm]
	Network & Loss & 5way1shot(CS) & Avg \# of Cls& 5way5shot(WS) \\
	\midrule
	FCGCN & MRL & $49.25\pm0.32\%$ & $3.64\pm0.03$ & $67.23\pm0.64\%$ \\
	Bi-LSTM& MRL & $48.57 \pm 0.75$ & $3.58\pm0.03$ & $66.61\pm0.98\%$ \\
	\midrule
	$\text{FL-GCN}_{(2)}$ & MSE & $51.18\pm0.17$ & $3.74\pm0.04$ & $67.05\pm0.34\%$\\
	\midrule
	$\text{FL-GCN}_{(1)}$ & MRL & $53.17\pm0.52\%$ & $3.93\pm0.06$ & $68.36\pm0.71\%$ \\
	\textbf{$\text{FL-GCN}_{(2)}$} & \textbf{MRL} & $\mathbf{54.00\pm0.64\%}$ & $\mathbf{4.05\pm0.04}$ & $\mathbf{69.14\pm0.78\%}$ \\
 	\bottomrule[0.3mm]
	\end{tabular}
	\label{tab:ablation}}
\end{table}

\vspace{-2mm}
\paragraph{Model struture} In Table.~\ref{tab:ablation}, we compare different model structures to verify the effectiveness of our proposed multi-components graph convolutional network. We first validate our proposed intra-and-inter message passing strategy by comparing with fully connected graph(FCGCN) and bi-directional LSTM(Bi-LSTM). The FCGCN propagates messages densely on the overall graph, which lacks the capacity to distinguish the contexts of the labeled and unlabeled set. The Bi-LSTM is sensitive to the order of data sequence. Our FL-GCN is superior to the other two baseline with a large margin.
Furthermore, experiments also indicate message passing with multiple iterations is helpful especially in the cold-start situation. 
\vspace{-2mm}
\paragraph{Training Loss}  As shown in Table.~\ref{tab:ablation}, we also investigate the effectiveness of the margin ranking loss(MRL) by comparing with the mean squared error loss(MSE), and the MRL can achieve better performance than the MSE in our few-shot learning task. Finally, we compute the average number of classes of these methods in the cold-start setting at the ending state, which implies our model is able to discriminate and find missing classes. 
\vspace{-2mm}
\paragraph{Anytime Performance} To measure the anytime performance, which means the model trained at each step to output the best possible prediction on the evaluation set, we plot the performance of different strategies at each step with a certain budget in Figure \ref{fig:anytime}. It's evident that our learned  policy performs consistently better than baseline methods at any step in both \textit{cold-start} and \textit{warm-start} situations.


%% file: data/conclusion.tex
\section{Conclusion}
To address the inefficiency of data sampling in previous few-shot learning, we have proposed a budget-aware few shot learning problem. Such tasks aim to jointly select training data and learn few-shot classifiers. To learn a generalized data selection policy, we develop a graph convolutional network with a novel message passing mechanism. To effectively learn the policy, we further propose to learn the policy in a supervised learning manner with pairwise margin ranking loss for the efficient learning. We demonstrate the efficacy of our approach by extensive experimental evaluations on the mini-ImageNet, tiered-ImageNet and Omniglot datasets. The quantitative results and ablation study clearly show that our method consistently improve performance of the budget-aware few-shot learning problem with efficiency and generalization ability.